\RequirePackage{fix-cm}
\documentclass[smallextended]{svjour3}       % onecolumn (second format)
\smartqed  % flush right qed marks, e.g. at end of proof
\usepackage{amssymb}
\usepackage{amsmath}
\usepackage{relsize}
\usepackage{graphicx}
\usepackage[font=small,skip=5pt]{caption}
\usepackage{multirow}
\usepackage{enumerate}
\usepackage{float}
\usepackage{subcaption}
\usepackage{textcomp}
\usepackage{csquotes}
\usepackage{array}
\usepackage{changes}
\usepackage{flexisym}
\DeclareMathOperator*{\argmax}{argmax}

\begin{document}

\title{An EM Based Probabilistic Two-Dimensional CCA with Application to Face Recognition
}
\titlerunning{Probabilistic 2DCCA with EM % if too long for running head
}
\author{Mehran Safayani \and Seyed Hashem Ahmadi \and Homayun Afrabandpey \and Abdolreza Mirzaei }

\institute{Department of Electrical and Computer Engineering, Isfahan University of Technology, Isfahan 84156-83111, IRAN \\
              Tel.: +98 31 33919063\\
              Fax: +98 31 33912451\\
              \email{safayani@cc.iut.ac.ir(Corresponding Author)\and hashem.ahmadi@ec.iut.ac.ir\and h.afraei@ec.iut.ac.ir \and mirzaei@cc.iut.ac.ir }
}

\date{Received: date / Accepted: date}

\maketitle

\begin{abstract}
Recently, two-dimensional canonical correlation analysis (2DCCA) has been successfully applied for image feature extraction. The method instead of concatenating the columns of the images to the one-dimensional vectors, directly works with two-dimensional image matrices. Although 2DCCA works well in different recognition tasks, it lacks a probabilistic interpretation. In this paper, we present a probabilistic framework for 2DCCA called probabilistic 2DCCA (P2DCCA) and an iterative EM based algorithm for optimizing the parameters. Experimental results on synthetic and real data demonstrate superior performance in loading factor estimation for P2DCCA compared to 2DCCA. For real data, three subsets of AR face database and also the UMIST face database confirm the robustness of the proposed algorithm in face recognition tasks with different illumination conditions, facial expressions, poses and occlusions.
\keywords{Canonical Correlation Analysis (CCA) \and Two-dimensional CCA \and Probabilistic Feature extraction \and Dimension Reduction \and Face recognition}
\end{abstract}

\section{Introduction}
\label{intro}
Although many real-world applications encounter high dimensional data, the most informative part of the data can be modeled in a low dimensional space. Moreover, processing high-dimensional data is a time consuming process and requires lots of resources. To tackle these problems, feature extraction has been used as a tool for finding a compact and meaningful data representation.\\
For single-mode source data, some subspace learning methods are conducted to learn more semantic description subspaces. Examples of these methods are principal component analysis (PCA) \cite{jolliffe2002principal} and linear discriminant analysis (LDA). However, for observations from two sources that share some mutual information, canonical correlation analysis (CCA) \cite{hotelling1936relations} is a very popular approach for dimensionality reduction. CCA seeks a lower-dimensional space where two sets of variables are maximally correlated after projecting on it. This technique is widely used in different fields of pattern recognition, computer vision, bioinformatics, etc. \cite{jia2012incremental,wang2014inferring,huang2014retracted}. In the CCA-based methods, it is necessary to vectorize 2D image matrices. Vectorization has three main drawbacks: (\textbf{I}) breaking the spatial structure of image data which may cause losing potentially useful structural information among column/rows \cite{ye2004gpca}, (\textbf{II}) leading to a high-dimensional vector space and small sample size problem which in turn makes it difficult to calculate the covariance matrices \cite{zhang20052d} and (\textbf{III}) causing the covariance matrices to be very large which in turn makes the eigen-decomposition of such large matrices very time-consuming.\\
To overcome these drawbacks, in 2007 two-dimensional CCA (2DCCA) was introduced by Lee and Choi \cite{lee2007two} which computes CCA directions based on 2D image matrices. The proposed 2DCCA overcomes the curse of dimensionality and significantly reduces the computational cost, by directly working with 2D images instead of reshaping them into 1D vectors. In \cite{lee2007two}, higher recognition accuracies were reported using 2DCCA compared to CCA using two face databases and the time complexity has been improved.\\
However, an associated probabilistic model for observed data was notably absent form these feature extraction methods. A probabilistic feature extraction algorithm could be intuitively appealing for so many reasons \cite{tipping1999probabilistic}. To bridge the gap, in 1999, Tipping and Bishop proposed probabilistic PCA \cite{tipping1999probabilistic} based on a latent variable model known as factor analysis (FA) \cite{thompson2004exploratory,browne1979maximum}. The proposed PPCA was then used as a framework for many other new formulations for PCA \cite{klami2008probabilistic,archambeau2006robust,archambeau2009sparse,zhao2012bilinear}.
Also, there have been some probabilistic models proposed for LDA \cite{ioffe2006probabilistic,kaski2003informative}. In 2005, Bach and Jordan \cite{bach2005probabilistic} also proposed a probabilistic interpretation of CCA and estimate the parameters of their proposed model using both maximum likelihood and expectation maximization. Recently, many inspiring research proceeded in the 1D CCA domain, including kernel based, semiparametric and nonparametric methods \cite{sarvestani2016ff,podosinnikova2016beyond,michaeli2016nonparametric}, but in the 2D CCA domain, we feel that more work is required. To bridge the gap, a probabilistic model of 2DCCA was introduced by Safayani et al. in 2011 \cite{safayani2011matrix}. They showed that the maximum likelihood estimation of parameters, leads to the two dimensional canonical correlation directions. However, they didn\textquotesingle t propose an EM based solution for their model. EM does not require the explicit eigen-decomposition of covariance matrices. Moreover, using EM it is possible to handle models with incomplete data such as mixture models where the cluster labels are the missing values \cite{wang2008probabilistic}.\\
In this paper, we present a probabilistic interpretation of 2DCCA, referred to as P2DCCA, together with an EM based solution to estimate the parameters of the model. The proposed model can handle the small sample size problem effectively.\\
The rest of the paper is organized as follows: Section \ref{SecII} briefly reviews some related algorithms such CCA, PCCA and 2DCCA which are necessary to understand how the proposed algorithms work. The proposed P2DCCA model is introduced in Section \ref{SecIII}. In Section \ref{SecIV}, some experiments on synthetic data and several face databases are given to evaluate performance of the proposed algorithm; finally, the paper is concluded in Section \ref{SecV}.
\section{Background}\label{SecII}

\subsection{Canonical Correlation Analysis (CCA)}
Imagine that we are given two sets of random vectors $t_{1}$ and $t_{2}$ where $t_{1,n}\in \textrm{R}^{D_{1}}$ and $t_{2,n}\in \textrm{R}^{D_{2}}$ for $n\in {1,2,...,N}$ are realizations of the corresponding random vectors, respectively. CCA seeks transformation vectors $w_{1}\in \textrm{R}^{D_{1}}$ and $w_{2}\in \textrm{R}^{D_{2}}$ such that correlation between $w_{1}^{T}t_{1}$ and $w_{2}^{T}t_{2}$ are maximized. The correlation between $w_{1}^{T}t_{1}$ and $w_{2}^{T}t_{2}$ can be formulated as
\begin{equation}\label{EQ1}
\rho = \frac{cov(w_{1}^{T}t_{1},w_{2}^{T}t_{2})}{\sqrt[2]{var(w_{1}^{T}t_{1})var(w_{2}^{T}t_{2})}}=\frac{w_{1}^{T}\Sigma_{12}w_{2}}{\sqrt[2]{(w_{1}^{T}\Sigma_{11}w_{1})(w_{2}^{T}\Sigma_{22}w_{2})}}
\end{equation}
\noindent
where $\Sigma_{ij}=\frac{1}{N}\sum_{n=1}^{N}(t_{i,n}-\mu_{i})(t_{j,n}-\mu_{j})^{T}$ for $i,j \in {1,2}$ is the cross-covariance matrix of $t_{1}$ and $t_{2}$ and $\mu_{i}=\frac{1}{N}\sum_{n=1}^{N}t_{i,n}$ for $i \in \{1,2\}$ denotes the mean vector of $t_{i}$.\\
\noindent
Then, the objective function for CCA can be written as:
\begin{flalign}\label{EQ2}
\argmax_{
\substack{w_{1},w_{2}}}
w_{1}^{T}\Sigma_{12}w_{2}&\\\nonumber
s.t.\;\; w_{1}^{T}\Sigma_{11}w_{2}=1\\\nonumber
\qquad w_{2}^{T}\Sigma_{22}w_{2}=1
\end{flalign}
\noindent
Optimizing such a constrained maximization problem with respect to $w_{1}$ and $w_{2}$ leads to the following generalized eigenvalue problem:
\begin{equation}\label{EQ3}
\begin{bmatrix}
0 & \Sigma_{12}\\
\Sigma_{21} & 0
\end{bmatrix}
\begin{bmatrix}
w_{1}\\
w_{2}
\end{bmatrix}
=\lambda \begin{bmatrix}
\Sigma_{11} & 0\\
0 & \Sigma_{22}
\end{bmatrix}
\begin{bmatrix}
w_{1}\\
w_{2}
\end{bmatrix}
\end{equation}
\noindent
By solving equation (\ref{EQ3}), $w_{1}$ and $w_{2}$ that maximize the correlation between the projected data can be found.

\subsection{Probabilistic CCA}

\begin{figure}[h!]
\begin{center}
\includegraphics[width=2.2in]{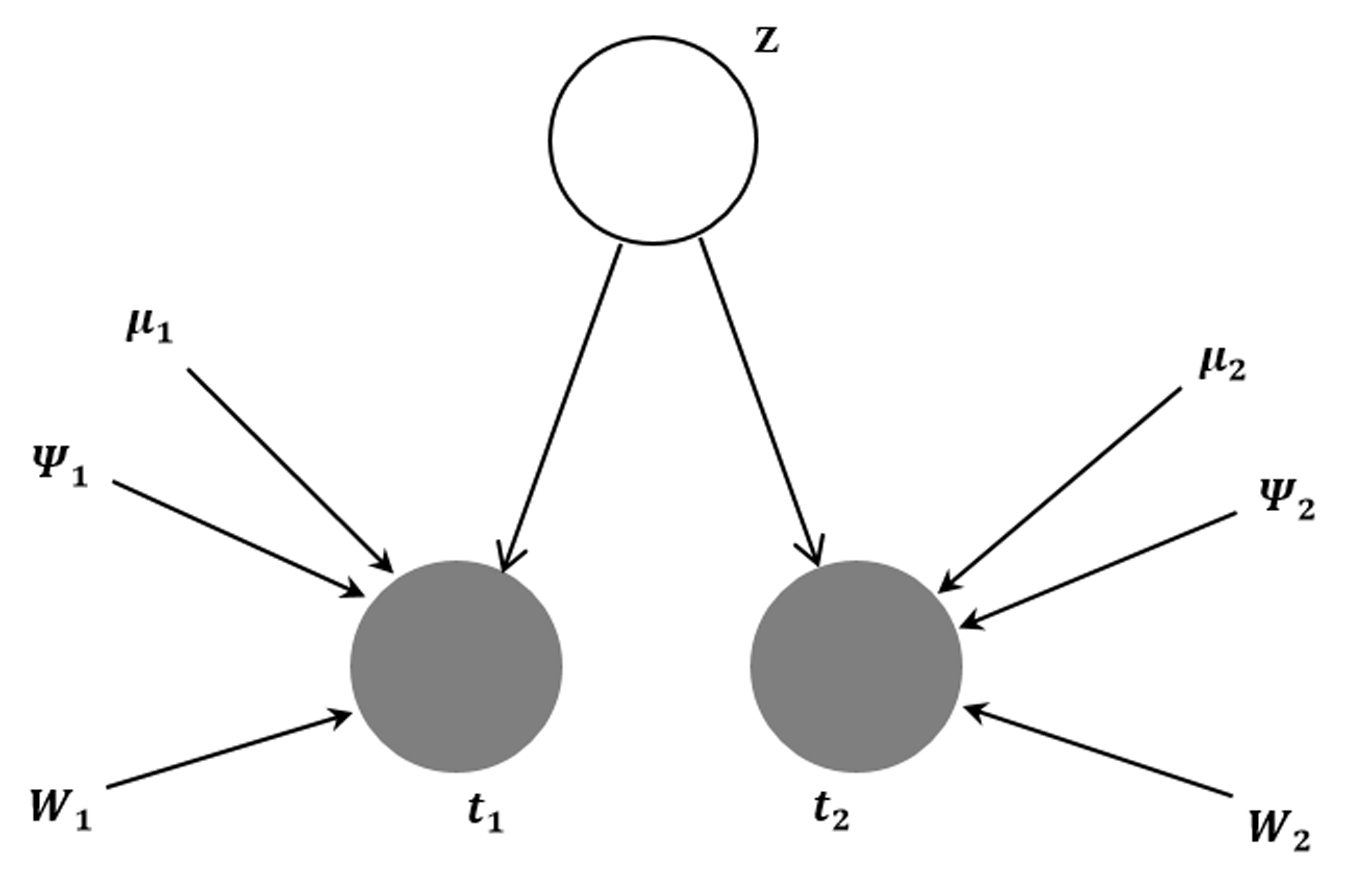}
\caption{Graphical model for probabilistic CCA}
\label{Fig1}
\end{center}
\end{figure}

\noindent
The generative model introduced by Bach and Jordan for CCA is as follows:
\begin{equation}\label{EQ4}
t_{i}=W_{i}z+\mu_{i}+\epsilon_{i} \qquad i\in \{1,2\}
\end{equation}
\noindent
In this model, $W_{i}\in \textmd{R}^{D_{i}\times d} \quad i\in \{1,2\}$ are linear projections that map two sets of high dimensional observed random vectors $t_{i}\in \textmd{R}^{D_{i}}, i\in \{1,2\}$ to a set of lower dimensional latent vectors $z\in \textmd{R}^{d}$. Here $\mu_{i}$ is the mean vector for $x_{i}$ and $\epsilon_{i}$ is the error term which is assumed to follow a multivariate Gaussian distributions with zero mean and inverse covariance matrix $\Psi_{i}$. Bach and Jordan proved that the maximum likelihood estimation for the parameters of this model would lead to the canonical directions. The maximum likelihood estimates of the projection matrices are given by:
\begin{equation}\label{EQ5}
\widehat{W}_{i}=\Sigma_{ii}U_{id}M_{i},\quad i\in \{1,2\}
\end{equation}
\noindent
where $\Sigma_{ii}$ is the sample covariance matrix, $U_{id}$ are the first $d$ canonical directions and $M_{i}\in \textmd{R}^{d\times d}$ where $i\in \{1,2\}$ are arbitrary matrices such that $M_{1}M_{2} = P_{d}$, where $P_{d}$ is the diagonal matrix of the first $d$ canonical correlations. Figure \ref{Fig1} shows a graphical representation of the model.

\subsection{Two-Dimensional CCA}

Two-dimensional CCA (2DCCA) was proposed to tackle the problem of vectorizing data in CCA. For each random matrix $T_{1}$ and $T_{2}$, 2DCCA introduces left transforms $u_{i}$ and right transforms $v_{i}$ where $i\in \{1,2\}$. After the projection, data would have the form $u_{i}^{T}T_{i}v_{i}$. 2DCCA finds these left and right transforms in a way to maximize the correlation between projected data. Therefore, the objective function of 2DCCA can be formulated as:
\begin{flalign}\label{EQ6}
&\argmax_{
\substack{u_{1},u_{2}},v_{1},v_{2}}
cov(u_{1}^{T}T_{1}v_{1},u_{2}^{T}T_{2}v_{2})&\\\nonumber
&s.t.\;\; var(u_{1}^{T}T_{1}v_{1})=1\\\nonumber
&\qquad var(u_{2}^{T}T_{2}v_{2})=1,&
\end{flalign}
\noindent
$u_{1}$ and $u_{2}$ can be obtained by solving the generalized eigenvalue problem (\ref{EQ7}) with fixed $v_{1}$ and $v_{2}$:
\begin{flalign}\label{EQ7}
\begin{bmatrix}
0 & \Sigma_{12}^{r}\\
\Sigma_{21}^{r} & 0
\end{bmatrix}\begin{bmatrix}
u_{1}\\
u_{2}
\end{bmatrix}=\lambda\begin{bmatrix}
\Sigma_{11}^{r} & 0\\
0 & \Sigma_{22}^{r}
\end{bmatrix}\begin{bmatrix}
u_{1}\\
u_{2}
\end{bmatrix}
\end{flalign}
In a similar way, given $u_{1}$ and $u_{2}$, right transforms $v_{1}$ and $v_{2}$ can be found by solving
\begin{flalign}\label{EQ8}
\begin{bmatrix}
0 & \Sigma_{12}^{l}\\
\Sigma_{21}^{l} & 0
\end{bmatrix}\begin{bmatrix}
v_{1}\\
v_{2}
\end{bmatrix}=\lambda\begin{bmatrix}
\Sigma_{11}^{l} & 0\\
0 & \Sigma_{22}^{l}
\end{bmatrix}\begin{bmatrix}
v_{1}\\
v_{2}
\end{bmatrix}
\end{flalign}
\noindent
where $\Sigma_{i,j}^{r}$, $i,j\in \{1,2\}$ is the cross-covariance matrix between $T_{i}$ and $T_{j}$ and $\Sigma_{ii}^{r}$, $i\in \{1,2\}$ is the auto-covariance matrix of $T_{i}$ defined as follows, respectively:
\begin{flalign}\label{EQ9}
\Sigma_{ij}^{r}=\frac{1}{N}\sum_{n=1}^{N}(T_{i,n}-\mu_{i})v_{i}v_{j}^{T}(T_{j,n}-\mu_{j})^{T},\qquad i,j\in\{1,2\}
\end{flalign}
\begin{flalign}\label{EQ10}
\Sigma_{ij}^{l}=\frac{1}{N}\sum_{n=1}^{N}(T_{i,n}-\mu_{i})u_{i}u_{j}^{T}(T_{j,n}-\mu_{j})^{T},\qquad i,j\in\{1,2\}
\end{flalign}
\noindent
where $T_{i,n}$ for $n\in\{1,...,N\}$ is the realization of the random matrix $T_{i}$ and $\mu_{i}=\frac{1}{N}\sum_{n=1}^{N}T_{i,n}$ is the corresponding matrix of mean values.\\
Left transforms ($l_{x}$ and $l_{y}$) and right transforms ($r_{x}$ and $r_{y}$) are obtained by iteratively solving equations (\ref{EQ7}) and (\ref{EQ8}), until convergence. The eigenvectors associated with $d_{1}$ largest eigenvalues in (\ref{EQ7}) determine left transform matrices $U_{1}$ and $U_{2}$ and the eigenvectors associated with $d_{2}$ largest eigenvalues in (\ref{EQ8}) determine right transform matrices $V_{1}$ and $V_{2}$. Using these transform matrices it is possible to project data from a high dimensional space to a new lower dimensional feature space.

\section{Probabilistic Two Dimensional CCA (P2DCCA)}\label{SecIII}

In this section, we propose probabilistic two-dimensional CCA and an EM-based solution for finding the parameters of the model. In our model, observed data are modeled as two-dimensional matrices as follows:
\begin{flalign}\label{EQ11}
T_{i}=U_{i}ZV_{i}^{T}+\mu_{i}+\Xi_{i} \quad i\in\{1,2\}
\end{flalign}
\noindent
where $T_{i}\in\textmd{R}^{m_{i}\times n_{i}}$ for $i\in \{1,2\}$ are observed matrices and $Z\in \textmd{R}^{m\textprime \times n\textprime}$ is the latent matrix. $U_{i} \in \textmd{R}^{m_{i}\times m\textprime}$, $V_{i} \in \textmd{R}^{n_{i}\times n\textprime}$ are projection matrices, $\mu_{i}$ is the mean matrix of the observed data and $\Xi_{i}$ is the residual matrix. Based on this definition, parameters of the model are $\{U_{i},V_{i}\}_{i=1}^{2}$ and the parameters of the distribution of $\Xi_{i}\textprime$.\\
Let $D_{i}={\{T_{i;n}\}}_{n=1}^{N}$ where $i\in \{1,2\}$ be a set containing $N$ observed data matrices and $\{Z_{n}\}_{n=1}^{N}$ be the corresponding latent variable set. Then the complete data would be $(T_{1,n},T_{2,n},Z_{n})$ and the log likelihood of the complete data can be written as
\begin{flalign}\label{EQ12}
L_{c}=\sum_{n=1}^{N}log \; p(T_{1,n},T_{2,n},Z_{n})
\end{flalign}
\noindent
To estimate the parameters, first we must calculate expectation of the log-likelihood and then take the derivative of the expected log-likelihood with respect to each parameter. Unfortunately there is no closed-form solution for computing the projection matrices $\{U_{i},V_{i}\}_{i=1}^{2}$ simultaneously. Inspired by \cite{tao2008bayesian}, a decoupled probabilistic model is employed to obtain projection matrices separately using an alternating optimization procedure. In such a model, we first assume that the value of one set of projection matrices, e.g. the right projections $\{V_{i}\}_{i=1}^{2}$, is known. Then observations are projected to the corresponding latent spaces. The projection procedure is a probabilistic one that introduced in section (\ref{SecPProjection}). By doing this, the left probabilistic model is defined as
\begin{flalign}\label{EQ13}
T_{i}^{l}=U_{i}Z^{l}+\Xi_{i}^{l}, \quad i=1,2
\end{flalign}
\noindent
where $Z^{l}$ is the left model latent matrix, $\mu_{i}^{l}$ is the mean matrix of the left projected observations and $\Xi_{i}^{l}$ is the noise source for left probabilistic model where columns of the noise matrix follow a normal distribution with zero mean and covariance matrix $\Psi_{i}^{l}$. By such definition, parameter set for the left probabilistic model would be $\Theta^{l}=\{U_{i},\Psi_{i}^{l}\}_{i=1}^{2}$ which can be estimated using expectation maximization procedure. The estimation procedure is explained later in this section.\\
In a similar procedure and parallel to the left probabilistic model, for the right probabilistic model, we assume that the left projection matrices, i.e. $\{U_{i}\}_{i=1}^{2}$, are known. Then observations are projected over the corresponding latent spaces,hence the right probabilistic model is defined as
\begin{flalign}\label{EQ14}
T_{i}^{r}=V_{i}Z^{r}+\Xi_{i}^{r}, \quad i=1,2
\end{flalign}
\noindent
 Similar to the left probabilistic model, $Z^{r}$, $\mu_{i}^{r}$ and $\Xi_{i}^{r}$ are defined for the right model where in this model the noise source have $N(0,\Psi_{i}^{r})$ distribution. The parameter set for the right probabilistic model would be $\Theta^{r}=\{V_{i},\Psi_{i}^{r}\}_{i=1}^{2}$. Using these definitions, the decoupled predictive density $p(T_{1},T_{2},Z)$ could be defined as
\begin{flalign}\label{EQ15}
p(T_{1},T_{2},Z)\propto p(T_{1}^{l},T_{2}^{l},Z^{l})p(T_{1}^{r},T_{2}^{r},Z^{r})
\end{flalign}
\noindent
Now we can rewrite the log likelihood of equation (\ref{EQ15}) as
\begin{flalign}\label{EQ16}
&L_{c}=\sum_{n=1}^{N}log(p(T_{1,n}^{l},T_{2,n}^{l},Z_{n}^{l})p(T_{1,n}^{r},T_{2,n}^{r},Z_{n}^{r}))&\\\nonumber
&\quad=\sum_{n=1}^{N}log \;p(T_{1,n}^{l},T_{2,n}^{l},Z_{n}^{l})+\sum_{n=1}^{N}log \;p(T_{1,n}^{r},T_{2,n}^{r},Z_{n}^{r}).&
\end{flalign}
To apply the EM algorithm to the decoupled probabilistic model, in E-step expectation of log likelihood for the left probabilistic model and the right probabilistic model is computed, separately. Then each of the expected log likelihood is maximized with respect to its parameters. In the following subsections we describe how to optimize left and right probabilistic model respectively.\\

\subsection{Optimizing the left probabilistic model}

Let $t_{i,j}^{l}$ be the $j^{th}$ column vector of $T_{i}^{l}\in \textmd{R}^{m_{i}\times n\textprime}$. By assuming columns of $T^{l}$ to be independent of each other, the distribution of $T_{i}^{l}$ is defined as
\begin{flalign}\label{EQ17}
p(T_{i}^{l})=\prod_{j=1}^{n\textprime}p(t_{i,j}^{l}), \qquad i \in \{1,2\}
\end{flalign}
\noindent
We also consider $z_{j}^{l} \in \textmd{R}^{m\textprime\times 1}$ as the $j^{th}$ column vector of $Z^{l}$ which has normal distribution of $N(0,I)$ and also in the same way $\mu_{i,j}^{l}$ is the $j^{th}$ column vector of $\mu_{j}^{l}$. Based on equation (\ref{EQ13}) and the distribution considered for $Z^{l}$ and $\Xi_{i}^{l}$, it can be concluded that
\begin{flalign}\label{EQ18}
p(t_{i,j}^{l})\sim N(\mu_{i,j}^{l},U_{i}U_{i}^{T}+\Psi_{i}^{l}), \qquad i=1,2
\end{flalign}
\noindent
Suppose $\tau_{n,j}^{l}= [(t_{1,n,j}^{l})^{T} \; (t_{2,n,j}^{l})^{T}]^{T} \in \textmd{R}^{(m_{1}+m_{2})\times 1}$, $V=[U_{1}^{T} \; U_{2}^{T}]^{T}\in \textmd{R}^{(m_{1}+m_{2})\times m\textprime}$, $m_{j}^{l}=[(\mu_{1,j}^{l})^{T} \; (\mu_{2,j}^{l})^{T}]^{T}\in \textmd{R}^{(m_{1}+m_{2})\times 1}$ and $\Psi^{l} = \left(
\begin{array}{cc}
\Psi_{1}^{l} & 0\\
0 & \Psi_{2}^{l}
\end{array}
\right)$
for the left probabilistic model, where $t_{i,n,j}^{l}$ refers to the $j^{th}$ column vector of $n^{th}$ image in the $i^{th}$ observation set where $i\in \{1,2\}$. Therefore, distributions of $p(\tau_{j}^{l})$ can be obtained as follows:
\begin{flalign}\label{EQ19}
p(\tau_{j}^{l})\sim N(m_{j}^{l},\Sigma^{l}),\qquad j\in[1,n\textprime]
\end{flalign}
\noindent
where $\Sigma^{l}=UU^{T}+\Psi^{l}$ and we assume $\Sigma^{l} > 0$.\\
Based on (\ref{EQ17}) we can write:
\begin{flalign}\label{EQ20}
p(T_{1,n}^{l},T_{2,n}^{l},Z_{n}^{l})=\prod_{j=1}^{n\textprime}p(\tau_{n,j}^{l},z_{n,j}^{l})=\prod_{j=1}^{n\textprime}p(\tau_{n,j}^{l}|z_{n,j}^{l})p(z_{n,j}^{l})
\end{flalign}
\noindent
To apply the EM algorithm to the decoupled probabilistic model, for each of the probabilistic models expectation of the log likelihood function is calculated in the E-step $E(L_{c}^{l})$ where the detail is given in the appendix, and then maximization step (M-step) is done by maximizing $E(L_{c}^{l})$ with respect to $V$ and $\Psi^{l}$. By doing so, the values of the parameters are estimated as
\begin{flalign}\label{EQ21}
U_{t+1}=\frac{\widetilde{\Sigma}^{l}(\Psi_{t}^{l})^{-1}U_{t}(M_{t}^{l})^{-1}}{(M_{t}^{l})^{-1}+(M_{t}^{l})^{-1}U_{t}^{l}(\Psi_{t}^{l})^{-1}\widetilde{\Sigma}^{l}(\Psi_{t}^{l})^{-1}U_{t}(M_{t}^{l})^{-1}}
\end{flalign}
\footnotesize
\begin{flalign}\label{EQ22}
\Psi_{t+1}^{l}=\left(
\begin{array}{cc}
(\widetilde{\Sigma}^{l}-\widetilde{\Sigma}^{l}(\Psi_{t}^{l})^{-1}U_{t}(M_{t}^{l})^{-1}U_{t+1}^{T})_{11} & 0\\
0 & (\widetilde{\Sigma}^{l}-\widetilde{\Sigma}^{l}(\Psi_{t}^{l})^{-1}U_{t}(M_{t}^{l})^{-1}U_{t+1}^{T})_{22}
\end{array}
\right)
\end{flalign}
\normalsize
\noindent
where $M^{l}=I+U^{T}(\Psi^{l})^{-1}U$ and $A_{t}$ shows the value of parameter $A$ in iteration $t$ and $\widetilde{\Sigma}^{l}$ is the sample covariance matrix of observed data for the left probabilistic model , i.e.
\begin{flalign}\label{EQ23}
\widetilde{\Sigma}^{l}=\frac{1}{N}\sum_{n=1}^{N}[(T_{1,n}^{l}-\mu_{1}^{l})^{T}\;(T_{2,n}^{l}-\mu_{2}^{l})^{T}]^{T}[(T_{1,n}^{l}-\mu_{1}^{l})^{T}\;(T_{2,n}^{l}-\mu_{2}^{l})^{T}]
\end{flalign}
\subsection{Optimizing the right probabilistic model}

In the manner similar to optimization of the left probabilistic model we have:
\begin{flalign}\label{EQ24}
p(T_{i}^{r})=\prod_{j=1}^{m\textprime}p(t_{i,j}^{r}), \qquad i \in \{1,2\}
\end{flalign}
\noindent
where $t_{i,j}^{r}$ is the $j^{th}$ column vector of $T_{i}^{r}\in \textmd{R}^{n_{i}\times m\textprime}$. Then $p(t_{i,j}^{r})$ is computed as:
\begin{flalign}\label{EQ25}
p(t_{i,j}^{r})\sim N(\mu_{i,j}^{r},V_{i}V_{i}^{T}+\Psi_{i}^{r}), \qquad i\in \{1,2\}
\end{flalign}
\noindent
Let $\tau_{n,j}^{r}= [(t_{1,n,j}^{r})^{T} \; (t_{2,n,j}^{r})^{T}]^{T} \in \textmd{R}^{(n_{1}+n_{2})\times 1}$, $V=[V_{1}^{T} \; V_{2}^{T}]^{T}\in \textmd{R}^{(n_{1}+n_{2})\times n\textprime}$, $m_{j}^{r}=[(\mu_{1,j}^{r})^{T} \; (\mu_{2,j}^{r})^{T}]^{T}\in \textmd{R}^{(n_{1}+n_{2})\times 1}$ and $\Psi^{r} = \left(
\begin{array}{cc}
\Psi_{1}^{r} & 0\\
0 & \Psi_{2}^{r}
\end{array}
\right)$, where $t_{i,n,j}^{r}$ refers to the $j^{th}$ column vector of $n^{th}$ image in the $i^{th}$ observation set. Then $p(\tau_{j}^{r})$ and $p(T_{1,n}^{r},T_{2,n}^{r},Z_{n}^{r})$ are obtained as follows:
\begin{flalign}\label{EQ26}
p(\tau_{j}^{r})\sim N(m_{j}^{r},\Sigma^{r}),\qquad j\in[1,m\textprime]
\end{flalign}
\begin{flalign}\label{EQ27}
p(T_{1,n}^{r},T_{2,n}^{r},Z_{n}^{r})=\prod_{j=1}^{m\textprime}p(\tau_{n,j}^{r},z_{n,j}^{r})=\prod_{j=1}^{m\textprime}p(\tau_{n,j}^{r}|z_{n,j}^{r})p(z_{n,j}^{r})
\end{flalign}
\noindent
where $\Sigma^{r}=VV^{T}+\Psi^{r}>0$. Given the details in appendix (A), after computing $E(L_{c}^{r})$ in the E-step, the parameters $V$ and $\Psi^{r}$ are computed by maximizing  the likelihood in the M-step. So, we have:
\begin{flalign}\label{EQ28}
V_{t+1}=\frac{\widetilde{\Sigma}^{r}(\Psi_{t}^{r})^{-1}V_{t}(M_{t}^{r})^{-1}}{(M_{t}^{r})^{-1}+(M_{t}^{r})^{-1}V_{t}^{r}(\Psi_{t}^{r})^{-1}\widetilde{\Sigma}^{r}(\Psi_{t}^{r})^{-1}V_{t}(M_{t}^{r})^{-1}}
\end{flalign}
\footnotesize
\begin{flalign}\label{EQ29}
\Psi_{t+1}^{r}=\left(
\begin{array}{cc}
(\widetilde{\Sigma}^{r}-\widetilde{\Sigma}^{r}(\Psi_{t}^{r})^{-1}V_{t}(M_{t}^{r})^{-1}V_{t+1}^{T})_{11} & 0\\
0 & (\widetilde{\Sigma}^{r}-\widetilde{\Sigma}^{r}(\Psi_{t}^{r})^{-1}V_{t}(M_{t}^{r})^{-1}V_{t+1}^{T})_{22}
\end{array}
\right)
\end{flalign}
\normalsize
\noindent
where $M^{r}=I+V^{T}(\Psi^{r})^{-1}V$ and $\widetilde{\Sigma}^{r}$ is computed as follows:
\begin{flalign}\label{EQ30}
\widetilde{\Sigma}^{r}=\frac{1}{N}\sum_{n=1}^{N}[(T_{1,n}^{r}-\mu_{1}^{r})^{T}\;(T_{2,n}^{r}-\mu_{2}^{r})^{T}]^{T}[(T_{1,n}^{r}-\mu_{1}^{r})^{T}\;(T_{2,n}^{r}-\mu_{2}^{r})^{T}]
\end{flalign}
\noindent

\subsection{Probabilistic projection and dimension reduction}\label{SecPProjection}
We can project the observation matrices into the latent space using the standard projection matrices ,i.e., $\{U_1,U_2,V_1,V_2\}$. However, as described in \cite{tipping1999probabilistic}, it is more natural to use probabilistic projections. In this regard, we represent each projected observation matrix, $T_i$, by mean of distribution of corresponding latent space, i.e., $E(Z|T_i)$. For the left model, it can be shown that
\begin{flalign}\label{EQpcca_1}
E(Z^l|T_1,T_2)=(M^{r})^{-1}\left[
\begin{array}{cc}
V_1^T & V_2^T\\
\end{array}
\right]\left[
\begin{array}{cc}
\Psi_1 & 0\\
0 & \Psi_2
\end{array}
\right]^{-1}\left[
\begin{array}{cc}
T_1-\mu_1\\
T_2-\mu_2
\end{array}
\right]
\end{flalign}

\noindent
$E(Z^l|T_1)$ and $E(Z^l|T_2)$ are obtained by marginalizing (\ref{EQpcca_1}) over $T_2$ and $T_1$ respectively. So we have
   \begin{flalign}\label{pcca_2}
E(Z^{l}|T_i)=(M^{r})^{-1}V_{i}^{T}(\Psi_{i}^{r})^{-1}(T_{i}-\mu_{i}), \qquad i\in \{1,2\}
\end{flalign}
 Similarly for the right model we have
   \begin{flalign}\label{pcca_3}
E(Z^{r}|T_i)=(M^{l})^{-1}U_{i}^{T}(\Psi_{i}^{l})^{-1}(T_{i}-\mu_{i}), \qquad i\in\{1,2\}
\end{flalign}
\noindent
The procedure for dimension reduction is given by sequential projection in left and right models as
\begin{flalign}\label{EQ31}
E(Z|T_{i})=(M^{l})^{-1}U_{i}^{T}(\Psi_{i}^{l})^{-1}(T_{i}-\mu_{i})((M^{r})^{-1}V_{i}^{T}(\Psi_{i}^r)^{-1})^{T}
\end{flalign}
\noindent
The P2DCCA algorithm is summarized in Figure \ref{fig:algorithm}. The proposed P2DCCA model benefits the ability to extend to other methods such as Mixtures of P2DCCA, Bayesian P2DCCA and also to robust P2DCCA.

\begin{figure}
    \begin{center}
    \begin{tabular}{|rlll|}
    \hline
    &&&\\
    1&\multicolumn{3}{l|}{\textbf{Input}: the matrices $T_{i,n}|_{n=1}^N\in\Re^{m_i\times n_i}$, $i\in\{1,2\}$}\\
    2&\multicolumn{3}{l|}{\textbf{Output}: Find $U_i\in \Re ^{m_i\times m^{'} }$, $V_i\in \Re^{n_i\times n^{'} }$ ,$\Psi^l$ and $\Psi^r$}\\
    3&\multicolumn{3}{l|}{\textbf{Initialize}: $V_{1,i}^l$ and $U_{1,i}^l$ using 2DCCA and initialize $\Psi_{1}^l$ and $\Psi_{1}^r$ with $I$ }\\
    4&\multicolumn{3}{l|}{\emph{for  t=1,...,}$T_{max}$}\\
    5&&\multicolumn{2}{l|}{$M^r=I+V_t^T(\Psi_t^r)^{-1}V_t$}\\
    6&&\multicolumn{2}{l|}{$T_{i,n}^{l}=(M^{r})^{-1}V_{t,i}^{T}(\Psi_{t,i}^{r})^{-1}(T_{i,n}-\mu_{i})$,  $i\in\{1,2\}$, $n=1,...,N$ }\\
   % 6&&\multicolumn{2}{l|}{assign $\psi_1^l$ and $\psi_2^l$ with small value}\\
    7&&\multicolumn{2}{l|}{\emph{Initialize} $\Psi_t^l$ randomly}\\
    8&&\multicolumn{2}{l|}{while $L_c^l$ not converge}\\
    9&&&{\emph{E-step:}}\\
    10&&&{\emph{Compute} $M^l=I+U_t^T(\Psi_t^l)^{-1}U_t$}\\
    11&&&{\emph{Compute} $E\{z_{n,j}^l\}=(M^l)^{-1}U_t^T(\Psi_t^l)^{-1}(\tau_{n,j}^l-m_{j}^{l})$}\\
    12&&&{\emph{Compute} $E\{(z_{n,j}^l)^Tz_{n,j}^l\}=(M^l)^{-1}+E\{z_{n,j}^l\}E\{z_{n,j}^l\}^T
$}\\
    13&&&{\emph{M-step:}}\\
    14&&&{$U_{t+1}=\frac{\Sigma_{n=1}^N\Sigma_{j=1}^{n^{'}}(\tau_{n,j}^l-m_{j}^{l})E\{z_{n,j}^l\}}{\Sigma_{n=1}^N\Sigma_{j=1}^{n^{'}}E\{z_{n,j}^l(z_{n,j}^l)^T\}}$}\\
    15&&&{$A^{l}=\frac{1}{n^{'}}\Sigma_{n=1}^N\Sigma_{j=1}^{n^{'}}(\tau_{n,j}^l-m_{j}^{l})(\tau_{n,j}^l-m_{j}^{l})^T+U_{t+1}E\{x_{n,j}^l\}(\tau_{n,j}^l-m_{j}^{l})
$}\\
16&&&{$\Psi_{t+1}^l=\left(
\begin{array}{cc}
A^{l}_{11} & 0\\
0 & A^{l}_{22}
\end{array}
\right)$}\\
    17&&\multicolumn{2}{l|}{End while}\\
    %%%%%%%%%%%%%%%%%%%%%%%%%%%%%%%%%%%%%%%%%%%%%%%%%%%%%%%%%% right projection
    18&&\multicolumn{2}{l|}{$M^l=I+U_{t}^T(\Psi_{t}^l)^{-1}U_{t}$}\\
    19&&\multicolumn{2}{l|}{$T_{i,n}^{r}=(M_{t}^{l})^{-1}U_{t,i}^{T}(\Psi_{t,i}^{l})^{-1}(T_{i,n}-\mu_{i})$,  $i\in\{1,2\}$, $n=1,...,N$ }\\
  %  17&&\multicolumn{2}{l|}{assign $\psi_r^l$ and $\psi_2^r$ with small value}\\
    20&&\multicolumn{2}{l|}{\emph{Initialize} $\Psi_t^r$ randomly}\\
    21&&\multicolumn{2}{l|}{while $L_c^r$ not converge}\\
    22&&&{\emph{E-step:}}\\
    23&&&{\emph{Compute} $M^r=I+V_t^T(\Psi^r)^{-1}V_t$}\\
    24&&&{\emph{Compute} $E\{z_{n,j}^r\}=(M^r)^{-1}V_t^T(\Psi^r)^{-1}(\tau_{n,j}^r-m_{j}^{r})$}\\
    25&&&{\emph{Compute} $E\{(z_{n,j}^r)^Tz_{n,j}^r\}=(M^r)^{-1}+E\{z_{n,j}^r\}E\{z_{n,j}^r\}^T
$}\\
    26&&&{\emph{M-step:}}\\
    27&&&{$V_{t+1}=\frac{\Sigma_{n=1}^N\Sigma_{j=1}^{m^{'}}(\tau_{n,j}^r-m_{j}^{r})E\{z_{n,j}^r\}}{\Sigma_{n=1}^N\Sigma_{j=1}^{m^{'}}E\{z_{n,j}^r(z_{n,j}^r)^T\}}
$}\\
    28&&&{$A^r=\frac{1}{m^{'}}\Sigma_{n=1}^N\Sigma_{j=1}^{m^{'}}(\tau_{n,j}^r-m_{j}^{r})(\tau_{n,j}^r-m_{j}^{r})^T+V_{t+1}E\{z_{n,j}^r\}(\tau_{n,j}^r-m_{j}^{r})
$}\\
    29&&&{$\Psi_{t+1}^r=\left(
\begin{array}{cc}
A^{r}_{11} & 0\\
0 & A^{r}_{22}
\end{array}
\right)$}\\

    30&&\multicolumn{2}{l|}{End while}\\
    31&\multicolumn{3}{l|}{\emph{end;}}\\
    &&&\\
    \hline
    \end{tabular}
    \end{center}
   \caption{Procedure of P2DCCA.}
   \label{fig:algorithm}

\end{figure}

\section{Experimental Results}\label{SecIV}

We evaluated our algorithm on both synthetic and real data. In synthetic data part we verified our implementation of P2DCCA algorithm by simple synthetic data and randomly generated projected matrices. we also compared our algorithm with 2DCCA method in projection matrices estimation . For real part evaluation, the proposed P2DCCA method was used for face recognition on two well-known face image databases (AR \cite{martinez1998ar} and UMIST \cite{graham1998characterising}). The AR database is divided into three subsets for evaluating the performance of the system in regard to different illumination, expression and occlusion conditions. The UMIST database is used to obtain the performance in dealing with pose variation.

\subsection{Experiments on synthetic data}
\begin{figure}[h!]
\begin{center}
\includegraphics[width=0.75\textwidth]{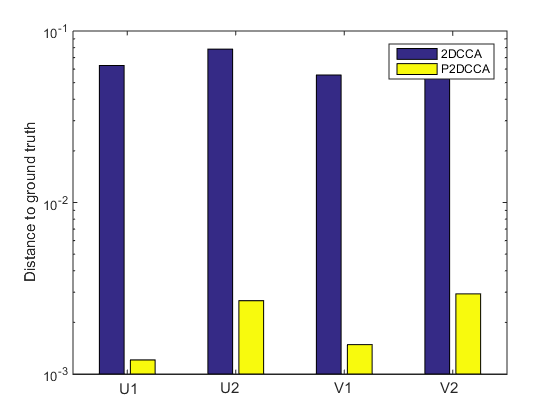}
\caption{Comparison of 2DCCA and P2DCCA in loading factors estimation on the synthetic data  with 1000 generated samples and $\sigma$ = 0.1}
\label{Fig3}
\end{center}
\end{figure}

\begin{figure}
\centering
\begin{subfigure}[b]{0.48\textwidth}
  \includegraphics[width=\textwidth, clip=true, trim=3cm 8cm 3cm 8cm]{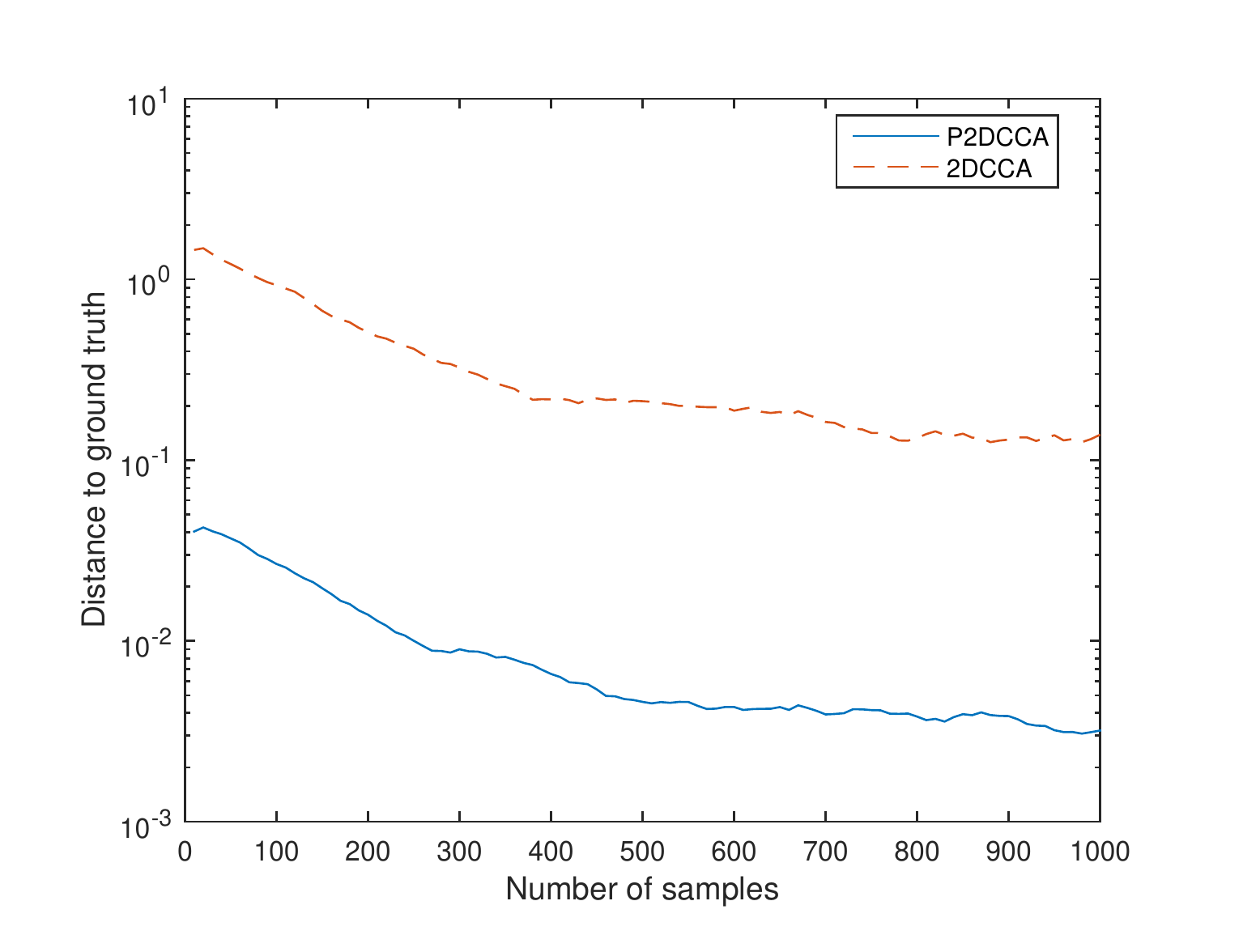}
  \caption{}
  \label{Fig3.A}
\end{subfigure}%
\hspace{0.03\textwidth}%
\begin{subfigure}[b]{0.48\textwidth}
  \includegraphics[width=\textwidth,clip=true, trim=3cm 8cm 3cm 8cm]{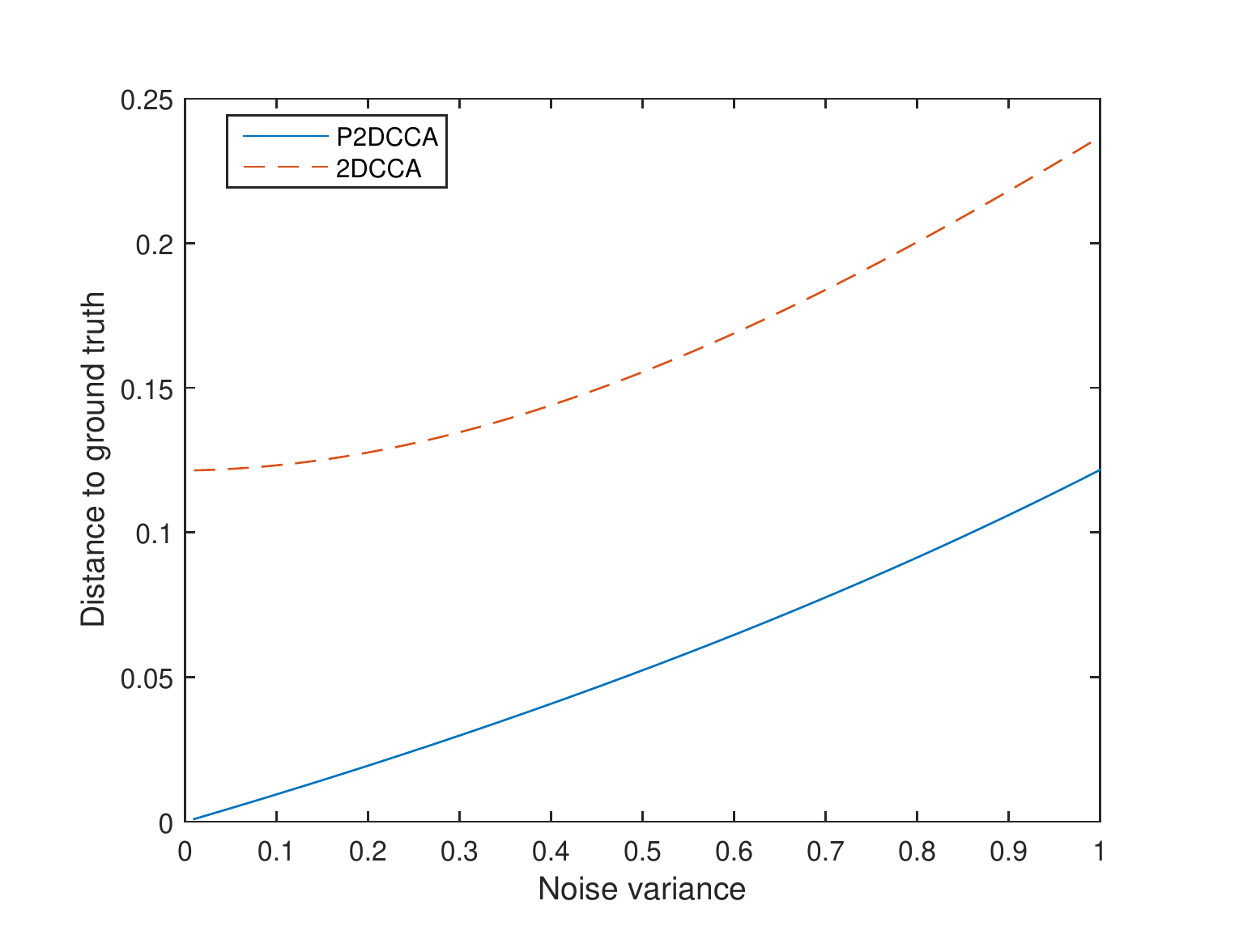}
  \caption{}
  \label{Fig3.B}
\end{subfigure}
\caption{Comparison of 2DCCA and P2DCCA in U1 loading factor estimation on the synthetic data}
\label{Sunth2}
\end{figure}

 In this section, we aim to verify our implementation of the proposed method with the simplest possible scenario. So we generate some synthetic data and projection matrices. Then estimate the projection matrices using our method and compare them with the true ones. We know that P2DCCA estimations are up to rotation, hence to simplify comparison we assume $Z$ to be $1 \times 1$ dimension. We set dimensions of $T_1$ and $T_2$ to $5 \times 5$. Then we generate 1000 samples of $Z$ from a normal distribution with zero mean and unit variance, also we randomly generate the elements of $U_i\in \Re ^{5}$, $V_i\in \Re ^{5}$, $i\in\{1,2\}$  using a uniform distribution in [0, 1] interval and consider them as the ground truth projection matrices, then $T_i$ will obtained using (\ref{EQ11}). In this equation, each element of the residual matrices sampled from a gaussian distribution with $0$ mean and $\sigma_i^2$ variance. Having the synthetic data we run P2DCCA algorithm as discussed in Figure \ref{fig:algorithm} and calculate $U_i$ and $V_i$. We also run 2DCCA and obtain the corresponding projection matrices. Then we compare obtained matrices by these two algorithms with the ground truth projection matrices. To cancel the scale factors we divide each transform by its norm before comparison. Euclidean distance is utilized to compare the normalized transforms. Figure \ref{Fig3} shows the results. It is obvious that in the worst-case the distance value becomes one and in the best-case it is zero. As it is depicted in this figure, P2DCCA estimation of $U_i$ and $V_i$ for $i\in\{1,2\}$ are much closer to the ground truth compared to the those obtained by 2DCCA. In this experiment we used 1000 generated samples and set $\sigma_i = 0.1$. To examine the effect of these selections, we repeat our experiment with different values of sample numbers and also noise variances. Figure \ref{Sunth2} demonstrates the results. As it can be observed from this figure the P2DCCA estimation in all cases is much closer to the ground truth compared to the 2DCCA method.

\noindent

\subsection{Experiments on the AR database}

\begin{figure}
\centering
\begin{subfigure}[b]{0.28\textwidth}
  \includegraphics[width=\textwidth]{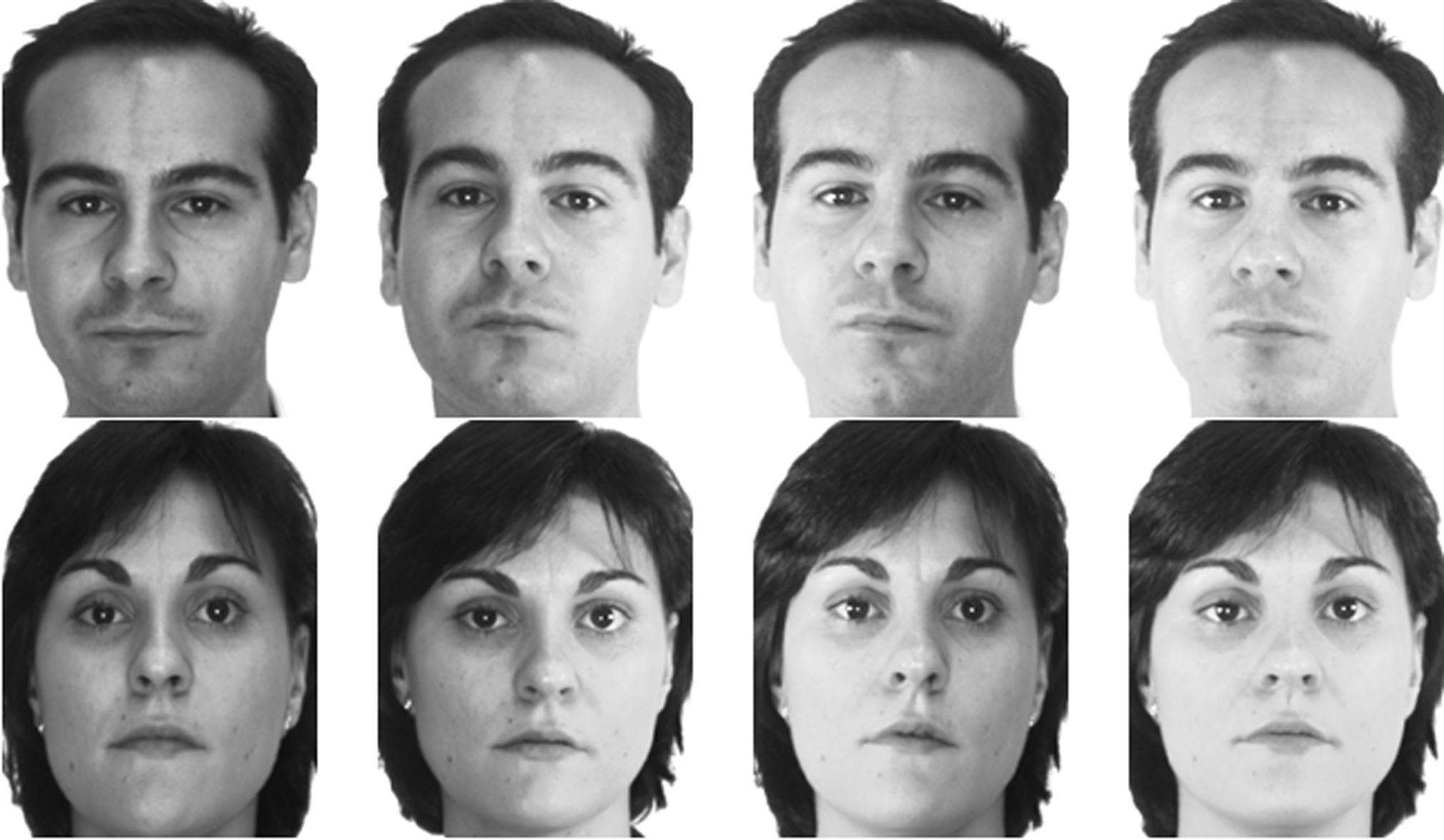}
  \caption{}
  \label{Fig4.A}
\end{subfigure}%
\hspace{0.03\textwidth}%
\begin{subfigure}[b]{0.28\textwidth}
  \includegraphics[width=\textwidth]{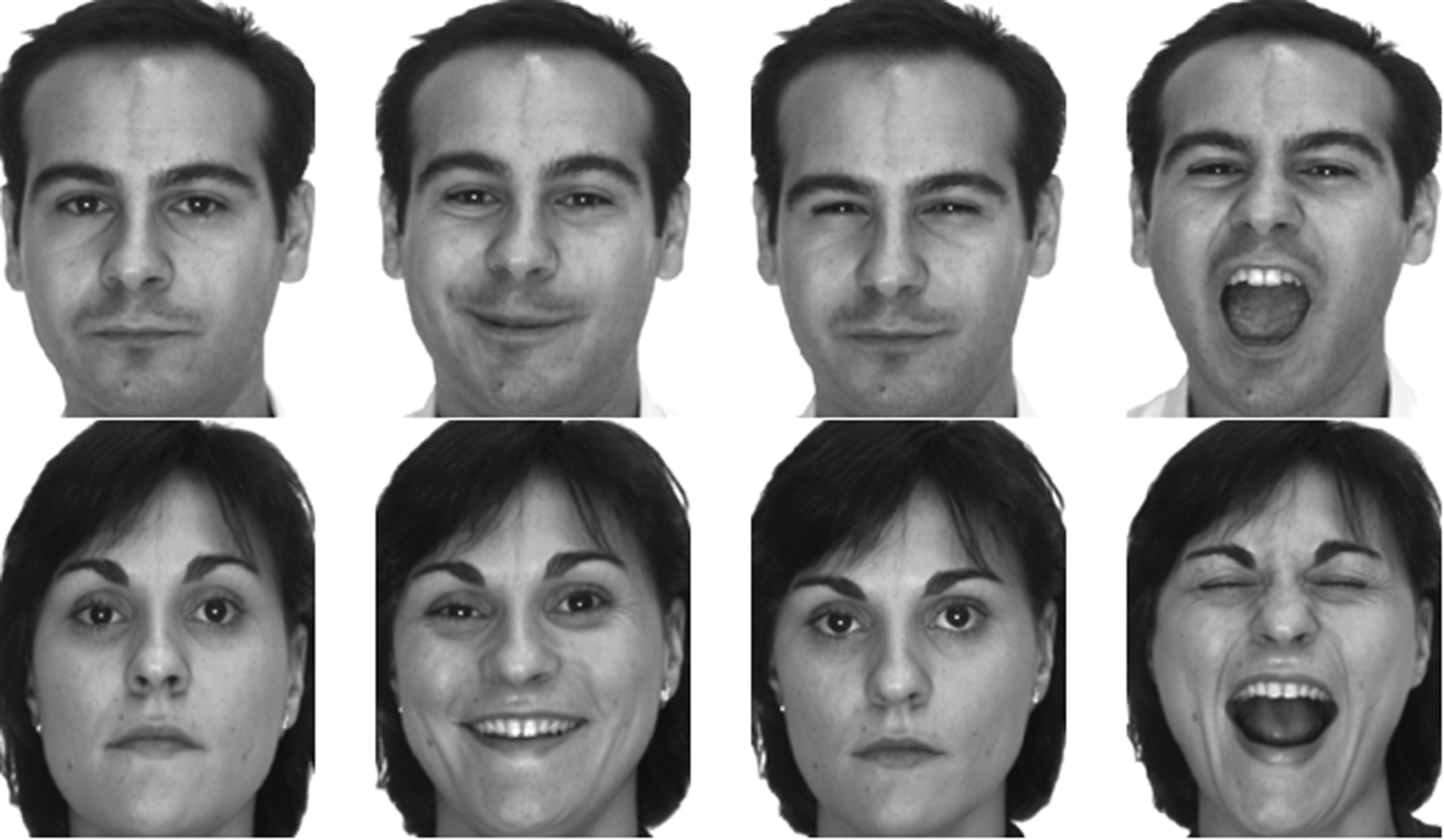}
  \caption{}
  \label{Fig4.B}
\end{subfigure}
\hspace{0.03\textwidth}
\begin{subfigure}[b]{0.28\textwidth}
  \includegraphics[width=\textwidth]{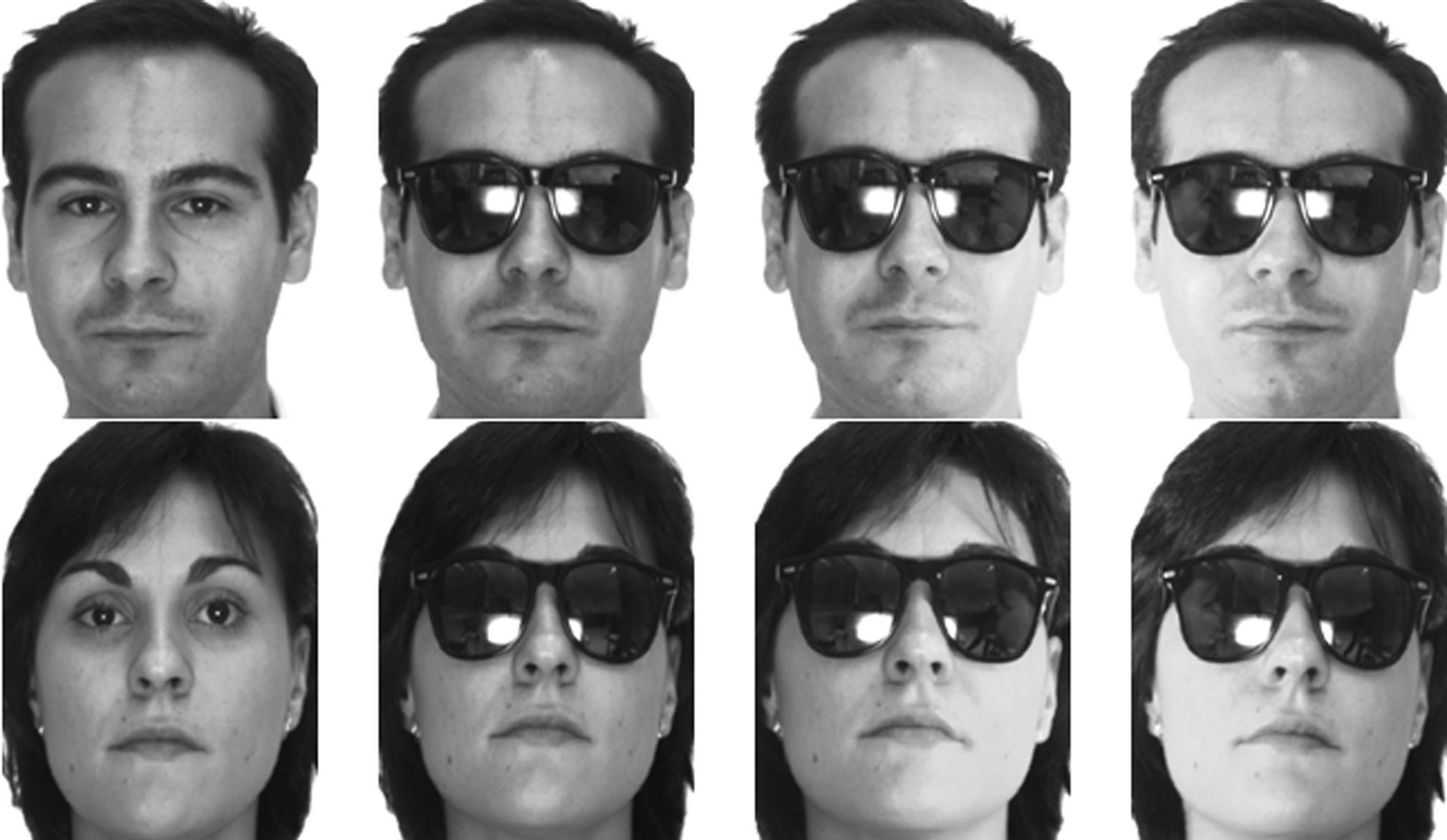}
  \caption{}
  \label{Fig4.C}
\end{subfigure}
\caption{Exemplary face images in (a). AR-1, (b). AR-2 and (c). AR-3}
\label{Fig4}
\end{figure}

The AR face database contains over 4,000 color face images including frontal views of faces with different facial expressions, illumination conditions and occlusions. For most individuals, there are two sessions of images which were taken in two different time periods. Each session contains 13 images. In our experiments, we used the first session, because some individuals do not have the second session of images. We collected 1310 face images of 131 people (72 male and 59 female). For each person, there are 10 different face images in our collected images: three with different illumination conditions; three with different expressions; three with occlusions and the remaining images are those with neutral expression and no occlusion which are known as reference images in our experiments. To examine the performance of the proposed methods in different conditions, we partitioned the collected images into three subsets known as AR-1, AR-2 and AR-3. For each individual, AR-1 contains four images, three of which are images with different lighting conditions and the remaining one is the reference image. AR-2 is used to test performance of the algorithms when there exist expression variation. AR-2 involves four images per individuals: three images have different expressions and the last one is the reference image. AR-3 is prepared to test performance in the presence of occlusion. Again, this subset contains four images per individuals where three images were taken with glasses and the last one is the reference image. Figure \ref{Fig4} shows exemplary face images of a man and a woman in AR-1, AR-2 and AR-3, respectively. Image are gray scaled, resized and then normalized to $50 \times 50$ pixels.
\noindent
 We compared the performance of the proposed method with a range of different supervised and unsupervised dimensionality reduction algorithms and different versions of them including PCA, LDA, CCA, PPCA, PCCA, 2DPCA, 2DLDA and 2DCCA. Both PCA and LDA based methods work with one set of data. Also LDA based methods are supervised while PCA and CCA based algorithms are unsupervised.
The task here is to investigate how well different algorithms can relate face images with varying illumination conditions, expressions and occlusion, in correspondence to the reference face images. The CCA, PCA, LDA, PPCA, PCCA, 2DCCA, 2DPCA, 2DLDA and P2DCCA are used to extract features from facial images and then a 1-NN classifier is employed for classification. Note that based on output type of each of the algorithms (vector or matrix), for 2DCCA, 2DPCA, 2DLDA and P2DCCA, Frobenius distance is used to calculate the distance between two feature matrices, while for CCA, PCA, LDA, PPCA and PCCA the common Euclidean distance measure is adopted. Furthermore, it should be noted here that since PCCA suffers the small sample size problem, implementing it using formulas introduced in \cite{bach2005probabilistic} caused the covariance matrices to be singular. To solve the problem, we did dimension reduction using PCA  before implementing the algorithm.\\

To evaluate the recognition accuracy, we used \enquote{three-fold cross-validation}. As it is evident, CCA based algorithms need two sets of images for training, where in this paper the training sets are called left training set and Right training set. To form the training sets, e.g., for AR-1, neutral images (images with no illumination) are considered as Left training set, while to form the Right training set, one of the three images of each individual with different illumination conditions is selected randomly. The other two images are considered as test images. This procedure is repeated for three times, where each time a different image among the three images is selected for the Right training set, while neutral images are always used to form the Left training set. To test the performance of each algorithm, all images of both right and left training sets are projected on the new feature spaces using their corresponding transforms. Also, each of the test images is projected on both feature spaces, so we have two projection for every test image. Then we calculate the distance between each of the two projected test images and projected training images. The label of the training image with the nearest projection to any of the two test image projection determine the final class of the test image. This procedure iterates until we find the final class for all images in the test set. These final classes are compared to the real classes of the images and the recognition accuracy of each algorithm is calculated. Finally, the average recognition rate of the three round experiments is recorded as the final recognition accuracy.\\

  Since PCA and LDA based algorithms work with one set of data, to have a fair comparison we used two images as training and the other two images as the test data, where neutral images are always in the training data together with one of the other images with different illuminations in each iterations. Again the process is repeated three times and the final accuracy is the average of the three runs.  Figure \ref{Fig5} shows the test process for P2DCCA.\\
\begin{figure}[h!]
\begin{center}
\includegraphics[width=2.8in]{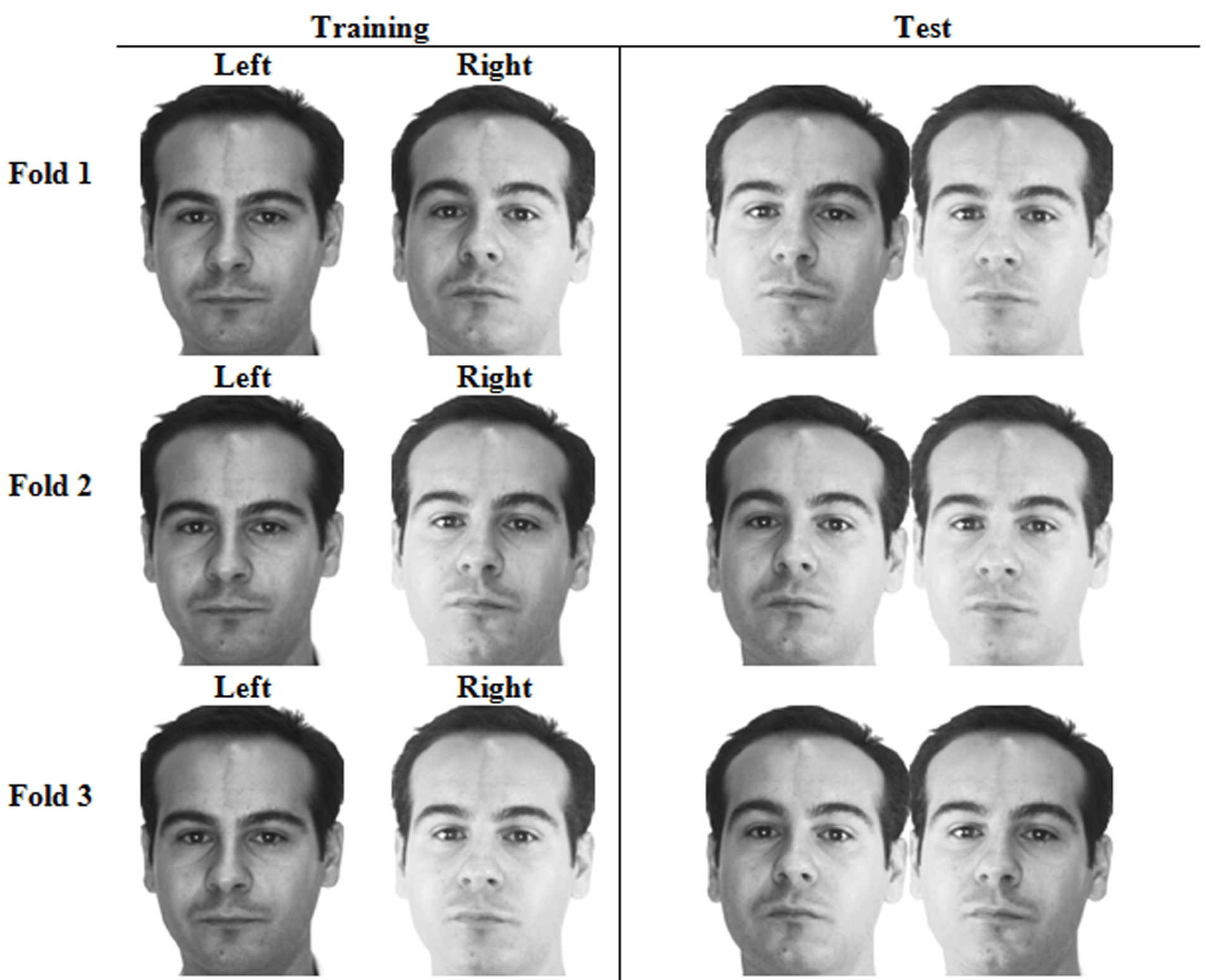}
\caption{A representation of training and testing images used in each fold for AR-1}
\label{Fig5}
\end{center}
\end{figure}
\noindent
Train and test procedure for AR2 and AR3 subsets are similar to AR1 and Table \ref{T1} through Table \ref{T3} demonstrate the recognition accuracy of evaluated algorithms for the experiments conducted on AR-1, AR-2 and AR-3, respectively. In these tables, $\textit{\textbf{d}}$ is the dimension of the reduced feature space. Note that output for two dimensional algorithms (2DCCA, 2DPCA, 2DLDA, P2DCCA and MP2DCCA) is of matrix type with dimension $\textit{\textbf{d}} \times \textit{\textbf{d}}$, while for CCA, PCA, LDA, PPCA and PCCA methods output is a vector of dimension $\textit{\textbf{d}}$. In these results we see that P2DCCA get the best performance among all tested methods.  We see about 10\% improvement of recognition rate for P2DCCA over 2DCCA in AR-1 and AR-3, and about 3\% improvement in AR-2.
\begin{table}[h!]
\caption{Comparison of the average recognition accuracy rates of the nine evaluated algorithms on AR-1 (\%)}
\begin{tabular}{c}
\includegraphics[width = 120 mm]{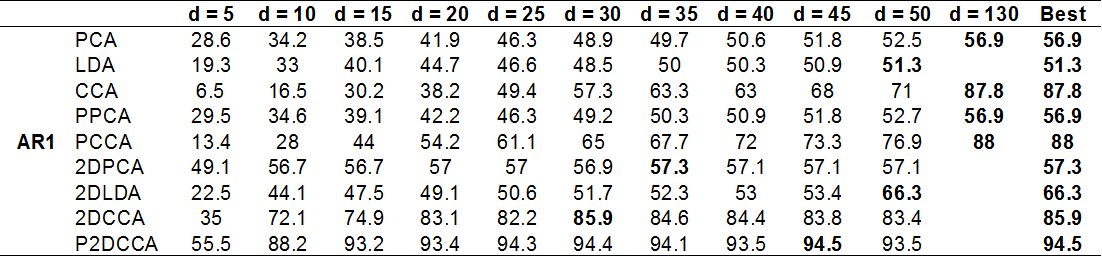}
\end{tabular}
\label{T1}
\end{table}
\begin{table}[h!]
\caption{Comparison of the average recognition accuracy rates of the nine evaluated algorithms on AR-2 (\%)}
\begin{tabular}{c}
\includegraphics[width = 120 mm]{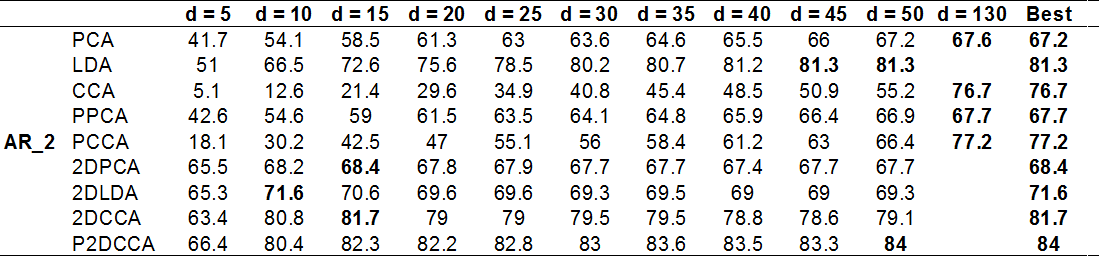}
\end{tabular}
\label{T2}
\end{table}
\begin{table}[h!]
\caption{Comparison of the average recognition accuracy rates of the nine evaluated algorithms on AR-3 (\%)}
\begin{tabular}{c}
\includegraphics[width = 120 mm]{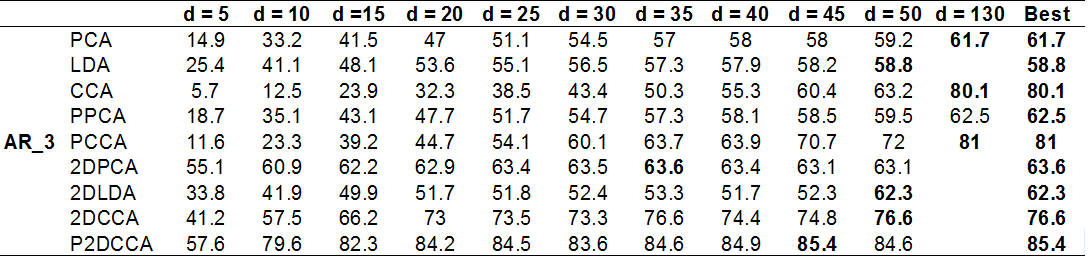}
\end{tabular}
\label{T3}
\end{table}
\noindent

\begin{figure}[h!]
\begin{center}
\includegraphics[width=0.75\textwidth, clip=true, trim=0cm 18cm 0cm 0cm]{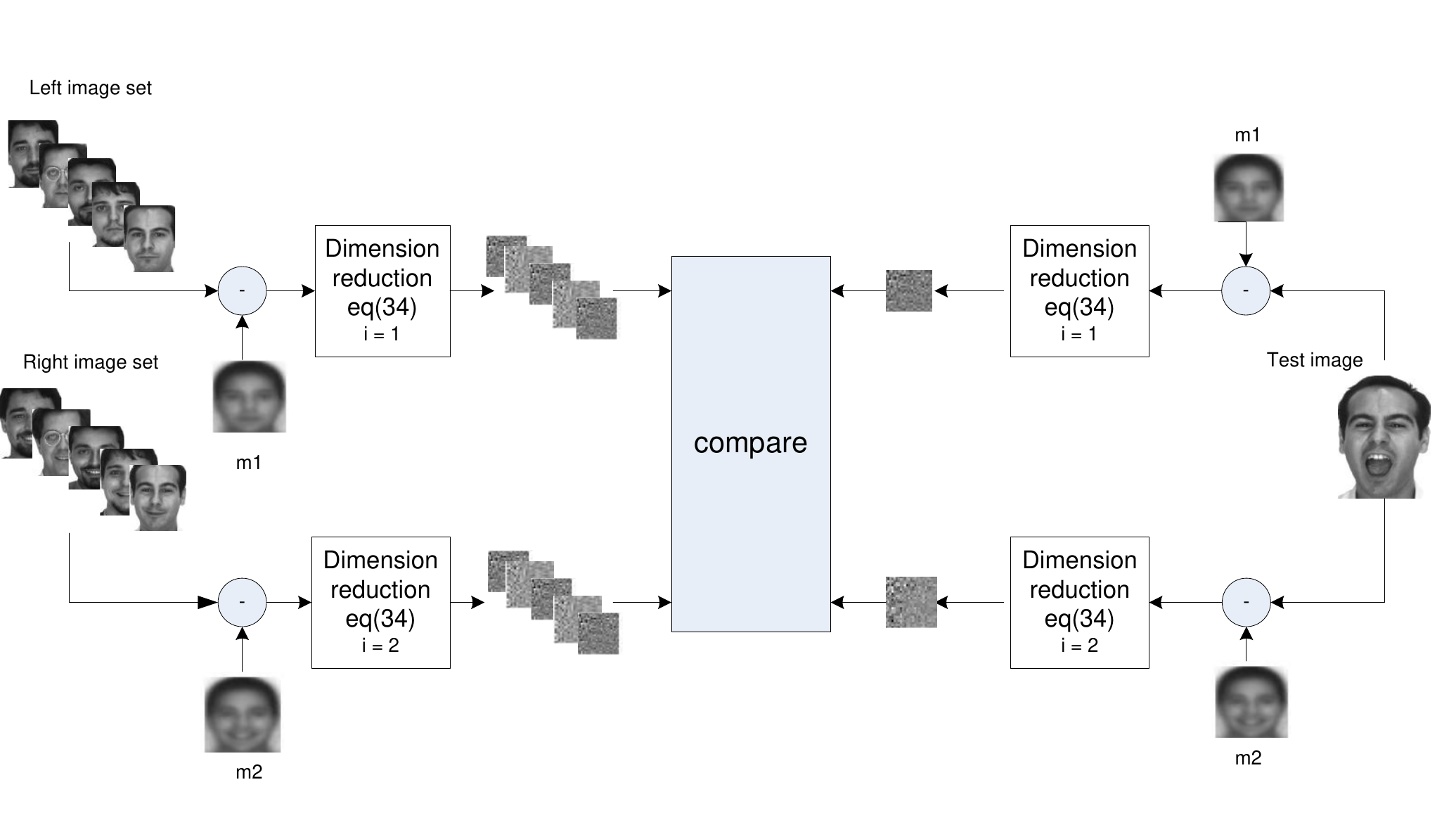}
\caption{Graphical representation of recognition process in P2DCCA}
\label{Fig5}
\end{center}
\end{figure}
\begin{figure}
\centering
\begin{subfigure}[b]{0.45\textwidth}
  \includegraphics[width=\textwidth]{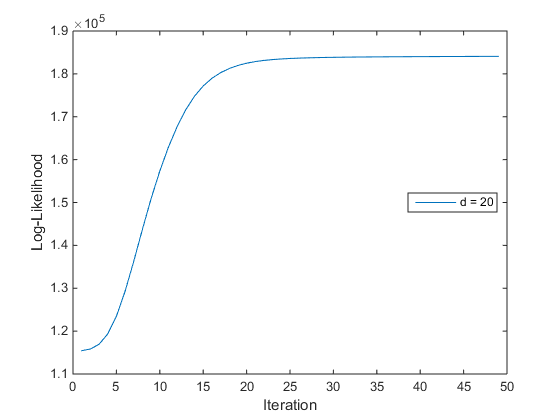}
  \caption{}
  \label{Fig3.A}
\end{subfigure}%
\hspace{0.02\textwidth}%
\begin{subfigure}[b]{0.45\textwidth}
  \includegraphics[width=\textwidth]{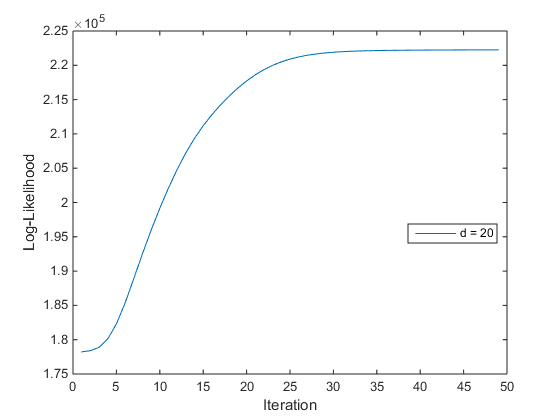}
  \caption{}
  \label{Fig3.B}
\end{subfigure}
\caption{Log-Likelihood values versus the number of iterations for (a) left probabilistic model and (b) Right probabilistic model of P2DCCA for d = 20}
\label{Fig6}
\end{figure}

\noindent
Figure \ref{Fig6} shows how the log-likelihoods of the left probabilistic model and the right probabilistic model of P2DCCA improve with each iteration. As it can be seen in the figure, both left and right models converge.

\begin{figure}
\begin{center}
\includegraphics[width=0.75\textwidth, clip=true, trim=3cm 8cm 3cm 8cm]{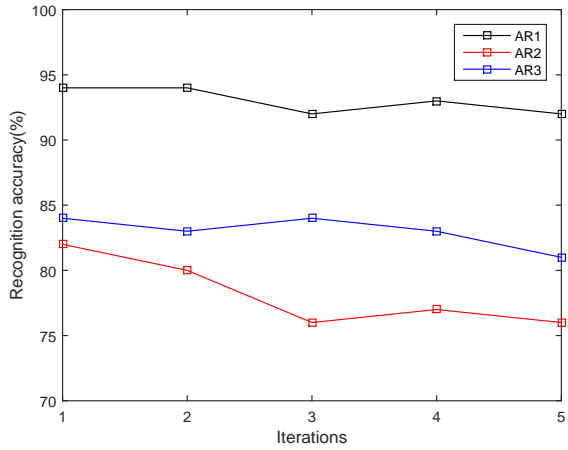}
\caption{Recognition rate of AR1, AR2 and AR3 for $Tmax > 1$}
\label{FigACC}
\end{center}
\end{figure}

It should be noted that it is very common that in the algorithms for optimizing row and column projections only one iteration with the iterative algorithm is performed\cite{confnipsYeJL04}. Therefore in all the experiments we use one iteration of the algorithm, i.e., $T_{max}=1$. This significantly reduces computational cost of the algorithm. We tried more iterations and got no significant improvement in the recognition rate. Figure \ref{FigACC} shows the results for 1 to 5 iterations for AR1, AR2 and AR3. This results support the idea of choosing $T_{max}=1$.

\subsection{Experiments on the UMIST database}
\begin{figure}[h!]
\begin{center}
\includegraphics[width=3.6in]{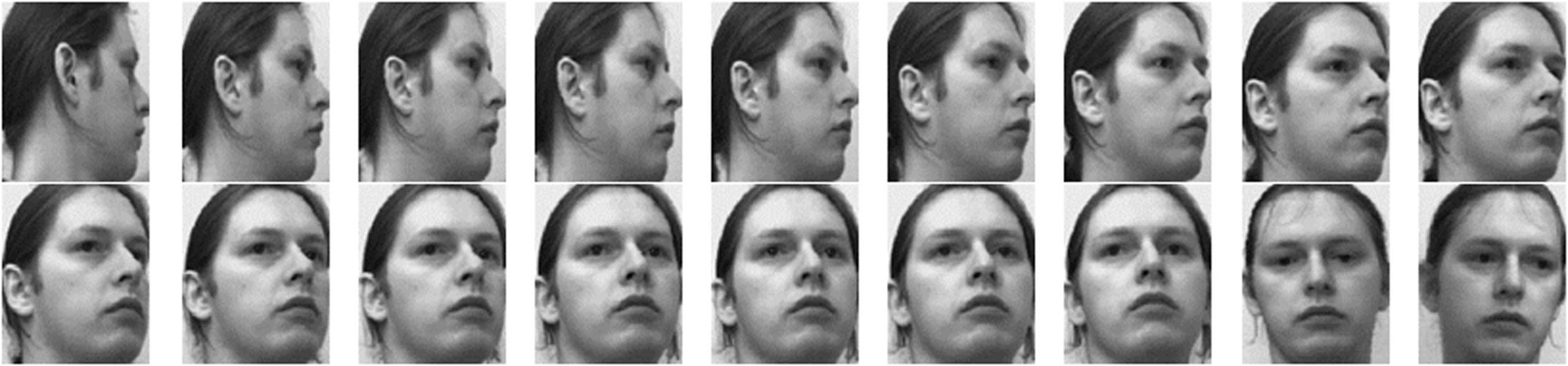}
\caption{Eighteen face examples of one subject from the UMIST database}
\label{Fig7}
\end{center}
\end{figure}

The UMIST face database also known as Sheffield face database \cite{UMIST} consists of 564 images of 20 subjects. Subjects have different races, sexes, and appearances. For each subject, there are images with different poses from profile to frontal view. Images have 256 grey levels with resolution of $220\times220$ pixels. In our experiment, 360 images with 18 samples per subject are used to examine performance of different algorithms when face orientation varies significantly. Figure \ref{Fig7} shows 18 images of one subject.

\noindent
We select frontal image as well as seven other randomly selected images for training set and the remaining images for the test set. In the training phase of CCA based methods, frontal image always is selected as the left training image and one of the seven other images as the right training image. 1-NN classifier is used for classification. This procedure is repeated for twenty times, and the average recognition rates of algorithms are reported. Table \ref{T4} shows the recognition accuracy of evaluated algorithms for the experiments conducted on UMIST. In this test while P2DCCA achieved slightly better performance compared to 2DCCA, it is not the best. In fact LDA achieved the best performance. In the AR test, LDA had 2 images per class for training, but here it has 8 images per class which leaded to best performance for this supervised method. Ignoring LDA, we see that P2DCCA performance is higher than other methods.\\

\begin{table}[h!]
\caption{Comparison of the average recognition accuracy rates of the nine evaluated algorithms on UMIST (\%)}
\begin{tabular}{c}
\includegraphics[width = 120 mm]{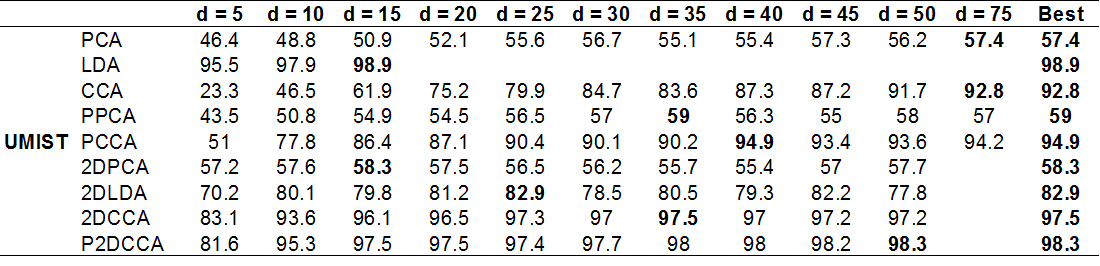}
\end{tabular}
\label{T4}
\end{table}
\noindent
Since the UMIST face dataset contains 20 subjects to be discriminated, the LDA features is limited to 19. To be able to compare the results of LDA with that of the other algorithms, we showed the results for $d=5$, $d=10$ and $d=15$ for LDA and larger values of $d$ for other algorithms.

\subsection{Evaluation of the Experimental Results}
The above experiments showed that the accuracy of P2DCCA is consistently better than other CCA based methods, i.e. CCA, PCCA and 2DCCA. But, a question sill remains: \enquote{Are these differences statistically significant?}. In this section we answered the question by evaluating the experimental results using independent-samples T-test (or independent t-test, for short). In this section and also the next section, we only considered CCA based algorithms including CCA, PCCA, 2DCCA and P2DCCA  since the goal of this paper is to compare the functionality of the newly proposed CCA based method with that of the other CCA based algorithms. The desired significance level is $0.05$ and the null hypothesis is that there is no significant difference between recognition rates of P2DCCA and CCA, PCCA and 2DCCA, respectively. We reject the null hypothesis whenever the resulted $\rho$-value becomes lower than $0.05$ and in this case the result can be considered statistically significant. It is necessary to note that to run the t-test in each dataset, for each algorithm we considered the highest recognition rate. Table \ref{T4} shows the $\rho$-value of the test. As can be seen from this table, P2DCCA significantly outperforms other algorithms and the null-hypothesis has been rejected in all cases.

\begin{table}[!ht]
\caption{The $\rho$-value associated with the null hypothesis: \enquote{no significant difference between recognition rates of P2DCCA and the corresponding algorithm}}
\begin{tabular}{c c c c}
\hline\noalign{\smallskip}
\textbf{Data Sets \textbackslash Algorithms} & \textbf{CCA} & \textbf{PCCA} & \textbf{2DCCA} \\
\noalign{\smallskip}\hline\noalign{\smallskip}
\textbf{AR-1} & 1.9e-06 & 9.5e-07 & 8.6e-10 \\
\textbf{AR-2} & 6.1e-12 & 5.7e-12 & 7.5e-05 \\
\textbf{AR-3} & 2.5e-05 & 4.8e-05 & 5.2e-06 \\
\textbf{UMIST} & 6.1e-05 & 3.5e-05 & 5.7e-03 \\
\noalign{\smallskip}\hline
\end{tabular}
\label{T4}
\end{table}
\vspace*{-\baselineskip}

\subsection{Computational Complexity}
This section compares the computational cost of the algorithms. To compare the time complexity of the algorithms, we consider input images of size $m \times m$ where we want to reduce their dimension to $d \times d$ in case of two-dimensional algorithms. For CCA and PCCA, vectorization caused the input data to have dimension $m^{2}\times 1$ and the output to have dimension $d \times 1$. However, for simplicity, $d$ is considered to be equal to $m$, i.e. $d=m$ in our analysis. Table \ref{T5} shows the computational complexity of the algorithms. In this table, $N$ is the number of random samples in the dataset. It should be noted that there are two types of iteration in the corresponding methods; one is the iteration necessary for the convergence of the EM part of the algorithm and the other is the iteration for alternating the optimization procedure between left and right model. We show the former by t and the latter by $r$ in the table. However, $r=1$ in our experiments.

\begin{table}[h!]
\caption{Time  complexity of algorithms}
\begin{tabular}{c}
\includegraphics[width = 90 mm]{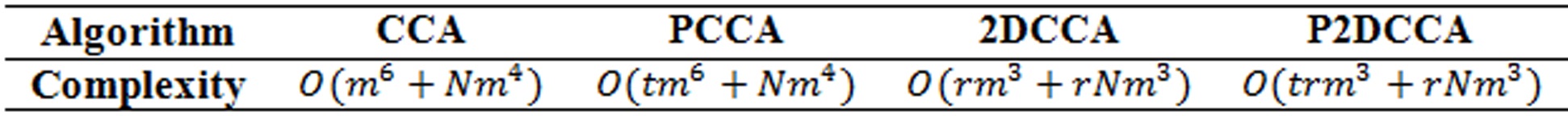}
\end{tabular}
\label{T5}
\end{table}
\vspace*{-\baselineskip}

\section{Conclusion}\label{SecV}
This paper proposed a probabilistic model for two dimensional CCA termed as P2DCCA together with an EM-based solution to estimate the parameters of the model. Experimental results demonstrated the functionality of the proposed method. The proposed P2DCCA has many advantages over 2DCCA where the most significant advantage is its ability to extend to a mixture of P2DCCA model. It may also be possible to develop a probabilistic Bayesian model for P2DCCA and gaining the benefits of a Bayesian model. These are our future works.

\bibliography{references}

\newpage
 \appendix
\section{}\label{Appendix}
As it is mentioned in Section \ref{SecIII}, each column of the latent matrix $Z$ has the distribution $N(0,I)$. Furthermore, based on (\ref{EQ13}) and (\ref{EQ14}) we can write:

\begin{flalign}
&p(\tau_{n,j}^{l}|z_{n,j}^{l})\sim N(Uz_{n,j}^{l}+m_{j}^{l},\Psi^{l})&
\end{flalign}
\vspace*{-\baselineskip}

\begin{flalign}
&p(\tau_{n,j}^{r}|z_{n,j}^{r})\sim N(Vz_{n,j}^{r}+m_{j}^{r},\Psi^{r}).&
\end{flalign}

\noindent
Now from (\ref{EQ20}) and (\ref{EQ27}) and the distribution of columns of Z, we have

\begin{flalign}
&p(T_{1,n}^{l},T_{2,n}^{l},Z_{n}^{l})=\prod_{j=1}^{n\textprime}[(2\pi)^{-\frac{m_{1}+m_{2}}{2}}|\Psi^{l}|^{-\frac{1}{2}}&\\\nonumber
&\qquad\qquad\qquad\qquad exp(-\frac{1}{2}(\tau_{n,j}^{l}-Uz_{n,j}^{l}-m_{j}^{l})^{T}(\Psi^{l})^{-1}(\tau_{n,j}^{l}-Uz_{n,j}^{l}-m_{j}^{l}))&\\\nonumber
&\qquad\qquad\qquad\qquad
(2\pi)^{-\frac{m\textprime}{2}}exp(-\frac{1}{2}(z_{n,j}^{l})^{T}(z_{n,j}^{l}))],&
\end{flalign}

\begin{flalign}
&p(T_{1,n}^{r},T_{2,n}^{r},Z_{n}^{r})=\prod_{j=1}^{m\textprime}[(2\pi)^{-\frac{n_{1}+n_{2}}{2}}|\Psi^{r}|^{-\frac{1}{2}}&\\\nonumber
&\qquad\qquad\qquad\qquad exp(-\frac{1}{2}(\tau_{n,j}^{r}-Vz_{n,j}^{r}-m_{j}^{r})^{T}(\Psi^{r})^{-1}(\tau_{n,j}^{r}-Vz_{n,j}^{r}-m_{j}^{r}))&\\\nonumber
&\qquad\qquad\qquad\qquad
(2\pi)^{-\frac{n\textprime}{2}}exp(-\frac{1}{2}(z_{n,j}^{r})^{T}(z_{n,j}^{r}))],&
\end{flalign}

\noindent
where $|A|$ denotes the determinant of matrix $A$.\\
In the E-step, expectation of log likelihood for each of the probabilistic models is calculated as:

\begin{flalign}
&E(L_{c}^{l})=\sum_{n=1}^{N}\sum_{j=1}^{n\textprime}[-\frac{m_{1}+m_{2}}{2}log(2\pi)-\frac{1}{2}log|\Psi^{l}|&\\\nonumber
&-\frac{1}{2}tr\{(\Psi^{l})^{-1}(\tau_{n,j}^{l})(\tau_{n,j}^{l})^{T}\}+tr\{(E[(z_{n,j}^{l})]-m_{j}^{l})^{T}U^{T}(\Psi^{l})^{-1}\tau_{n,j}^{l}\}&\\\nonumber
&-\frac{1}{2}tr\{(E[z_{n,j}^{l}]-m_{j}^{l})^{T}U^{T}(\Psi^{l})^{-1}U(E[z_{n,j}^{l}]-m_{j}^{l})\}-\frac{m\textprime}{2}log(2\pi)&\\\nonumber
&-\frac{1}{2}tr\{E[(z_{n,j}^{l})(z_{n,j}^{l})^{T}]\}]&
\end{flalign}

\begin{flalign}
&E(L_{c}^{r})=\sum_{n=1}^{N}\sum_{j=1}^{m\textprime}[-\frac{n_{1}+n_{2}}{2}log(2\pi)-\frac{1}{2}log|\Psi^{r}|&\\\nonumber
&-\frac{1}{2}tr\{(\Psi^{r})^{-1}(\tau_{n,j}^{r})(\tau_{n,j}^{r})^{T}\}+tr\{(E[(z_{n,j}^{r})]-m_{j}^{r})^{T}V^{T}(\Psi^{r})^{-1}\tau_{n,j}^{r}\}&\\\nonumber
&-\frac{1}{2}tr\{(E[z_{n,j}^{r}]-m_{j}^{r})^{T}V^{T}(\Psi^{r})^{-1}V(E[z_{n,j}^{r}]-m_{j}^{r})\}-\frac{n\textprime}{2}log(2\pi)&\\\nonumber
&-\frac{1}{2}tr\{E[(z_{n,j}^{r})(z_{n,j}^{r})^{T}]\}]&
\end{flalign}

\begin{flalign}
&E[z_{n,j}^{l}]=(M^{l})^{-1}U^{T}(\Psi^{l})^{-1}(\tau_{n,j}^{l}-m_{j}^{l})&
\end{flalign}
\begin{flalign}
&E[(z_{n,j}^{l})(z_{n,j}^{l})^{T}]=(M^{l})^{-1}+E[(z_{n,j}^{l})]E[(z_{n,j}^{l})]^T&
\end{flalign}
\begin{flalign}
&E[z_{n,j}^{r}]=(M^{r})^{-1}V^{T}(\Psi^{r})^{-1}(\tau_{n,j}^{r}-m_{j}^{r})&
\end{flalign}
\begin{flalign}
&E[(z_{n,j}^{r})(z_{n,j}^{r})^{T}]=(M^{r})^{-1}+E[(z_{n,j}^{r})]E[(z_{n,j}^{r})]^T.&
\end{flalign}

\noindent
By obtaining formulas of $E(L_{c}^{l})$ and $E(L_{c}^{r})$, the M-step is done by derivation of each expected log-likelihoods.

\end{document}